\documentclass[11pt]{article}
\usepackage{graphicx,caption}
\usepackage{coling2020}
\usepackage{times}
\usepackage{float}
\usepackage{url}
\usepackage{tabularx}
\usepackage{multirow}
\usepackage{booktabs}
\usepackage{tabu}
\usepackage{latexsym}
\usepackage{amsmath}

\usepackage{tikz}
\usetikzlibrary{calc,trees,positioning,arrows,chains,shapes.geometric,%
    decorations.pathreplacing,decorations.pathmorphing,shapes,%
    matrix,shapes.symbols,shadows,shapes.geometric}
    
\tikzset{
>=stealth',
  punktchain/.style={
    rectangle, 
    rounded corners, 
    draw=black, very thick,
    text width=10em, 
    minimum height=3em, 
    text centered, 
    on chain},
  line/.style={draw, thick, <-},
  element/.style={
    tape,
    top color=white,
    bottom color=blue!50!black!60!,
    minimum width=8em,
    draw=blue!40!black!90, very thick,
    text width=10em, 
    minimum height=3.5em, 
    text centered, 
    on chain},
  every join/.style={->, thick,shorten >=1pt},
  decoration={brace},
  tuborg/.style={decorate},
  tubnode/.style={midway, right=2pt},
startend1/.style={
    rectangle, 
    rounded corners=15pt, 
    text width=3cm, 
    minimum height=1cm,
    align=center, 
    line width=2pt,
    draw=white,
    font=\color{black}\sffamily, 
    fill=white
  }}

\usepackage{gb4e}
\newcommand{\corpusname}[0]{\textsc{Knowref-60k}}
\noautomath 
\newcolumntype{L}[1]{>{\raggedright\let\newline\\\arraybackslash\hspace{0pt}}m{#1}}
\newcolumntype{C}[1]{>{\centering\let\newline\\\arraybackslash\hspace{0pt}}m{#1}}
\newcolumntype{R}[1]{>{\raggedleft\let\newline\\\arraybackslash\hspace{0pt}}m{#1}}

\title{An Analysis of Dataset Overlap on Winograd-Style Tasks}

\author{\\\textbf{Ali Emami} \\
  Mila/McGill University \\
  {\tt ali.emami@mail.mcgill.ca } \\ \\
  \textbf{Adam Trischler, Kaheer Suleman} \\
  Microsoft Research Montreal \\
  {\tt \{adam.trischler, kasulema\}@microsoft.com}\\ \\
  \textbf{Jackie Chi Kit Cheung} \\
  Mila/McGill University \\
  {\tt jcheung@cs.mcgill.ca}}

\date{}

\begin{document}

\maketitle
\vspace{2cm}
\begin{abstract}

The Winograd Schema Challenge (WSC) and variants inspired by it have become important benchmarks for common-sense reasoning (CSR). Model performance on the WSC has quickly progressed from chance-level to near-human using neural language models trained on massive corpora. In this paper, we analyze the effects of varying degrees of overlap between these training corpora and the test instances in WSC-style tasks. We find that a large number of test instances overlap considerably with the corpora on which state-of-the-art models are (pre)trained, and that a significant drop in classification accuracy occurs when we evaluate models on instances with minimal overlap. Based on these results, we develop the \corpusname{} dataset, which consists of over 60k pronoun disambiguation problems scraped from web data. \corpusname{} is the largest corpus to date for WSC-style common-sense reasoning and exhibits a significantly lower proportion of overlaps with current pretraining corpora.

\end{abstract}

\section{Introduction}

    The original purpose of the Winograd Schema Challenge was to serve as an alternative Turing test to evaluate an automatic system's capacity for common-sense inference \cite{levesque2011winograd}. As an example:
\begin{exe}
\ex \begin{xlist} 
    \ex \label{ex-1a} Jim \underline{yelled at} Kevin because \textit{he} was so upset. (Answer: Jim)
    \ex \label{ex-1b} Jim \underline{comforted} Kevin because \textit{he} was so upset. (Answer: Kevin)
    \end{xlist}
    \label{exe:representation}
\end{exe}
For a number of years, models struggled to exceed chance-level performance \cite{kruengkrai2014example,sharma2015towards,peng2015solving,liu2016probabilistic}. The WSC task is carefully controlled, such that heuristics involving syntactic  and  semantic  cues  were ineffective, and the common-sense knowledge required to correctly resolve its test instances make it particularly difficult for statistical systems to model. More recently, however, the advent of deep bidirectional transformers (e.g., BERT \cite{devlin2018bert}, RoBERTa \cite{liu2019roberta}) pretrained on massive amounts of data has led to near-human-level performance \cite{kocijan-etal-2019-surprisingly,ye2019align,ruan2019exploring}. Various works have lately re-examined the challenges of the WSC, leading to the proposal of more difficult, larger variants, data and model debiasing methods, and evaluation protocols that clarify which types of instances models excel on and which they struggle with \cite{trichelair2018evaluation,emami-etal-2019-KnowRef,sakaguchi2019winogrande,abdou2020sensitivity}. However, little attention has been paid to studying the effects and influence of pretraining data points. 
While recent work has included some analysis on the effects of 13-gram overlaps between pretraining and test instances for the WSC \cite{brown2020language}, a deeper look into how the \textit{degree} of overlap (and how this can be defined) affects language models' performance is critical to revealing models' reasoning and inference functions.  For example, in \ref{ex-1b}), useful knowledge instances in the pretraining corpora may occur as:

\begin{exe}
\ex George comforted Melissa because she was very upset. (high overlap)
\ex And he comforted me because I was so upset by the whole event.  (lower overlap)
\end{exe}

Studying how models make use of these training instances, in correlation with their overlap with test instances, can provide insight on the roles and downstream influence of exact duplicates (which may be useful if memorized) or highly relevant but distinctly expressed knowledge (useful by retrieval and analogy). In turn, this insight could be used to improve training approaches for models meant to exhibit common-sense reasoning.

\paragraph{Contributions:} In this work, we address the above issues in CSR modeling by devising a mechanism to score train-test overlap according to a schematization based on BM25, a popular information retrieval function for text matching \cite{Amati2009}. We use the mechanism to sub-divide test-set instances according to these overlaps. We find that a significant drop in classification accuracy occurs when models are evaluated on the subset with no overlap (we see drops between 3\% and 10\%, depending on the model, test set, and degree of overlap). Based on this result, we develop the \corpusname{} dataset, consisting of 64,301 difficult pronoun disambiguation problems. It is the largest corpus to date for WSC-style common sense reasoning and exhibits a significantly lower proportion of overlaps with current pretraining corpora.\footnote{The corpus, the code to scrape the sentences from the source texts, as well as the code to reproduce all experimental results will be publicly available at https://github.com/aemami1/KnowRef60k.}

\section{Related Work}

Previous work on the difficulty of instances in the WSC and its variants includes the study by \newcite{trichelair2018evaluation}, who classified data points into various meaningful subsets.
They showed that the success of a then-state-of-the-art LM ensemble \cite{trinh2018simple} resulted mainly from improvements on simpler ``associative'' instances.
Similarly, experiments by \newcite{abdou2020sensitivity} show that models are sensitive to linguistic perturbations of Winograd-style examples. New datasets have been proposed to circumvent issues of unintentionally easy test instances, including Winogrande \cite{sakaguchi2019winogrande}, a scaled WSC-variant debiased against RoBERTa, and KnowRef \cite{emami-etal-2019-KnowRef}, which consists of naturally occurring sentences that are free of WSC-specific stylistic quirks. 

Given the recent popularity of large, internet-scale datasets for pretraining neural language models, there is an increasing concern that test instances in a downstream task may inadvertently appear in the pretraining corpus. This is a form of data contamination. One of the earliest works that trained a language model on Common Crawl data identified and removed a training documents that overlapped with one of their evaluation datasets \cite{trinh2018simple}. Other work, such as GPT-2 \cite{radford2019language} and GPT-3 \cite{brown2020language}, conducted post-hoc overlap analysis on CSR benchmarks based on a conservative threshold for contamination---specifically, instance pairs that have a 13-gram overlap. They found the effects of this 13-gram contamination to be negligible. On the other hand, a recent work in computer vision found a significant effect of \emph{near}-duplicates on test performance in an important benchmark, leading to the proposal of a duplicate-free and demonstrably more difficult dataset \cite{barz2020we}. To our knowledge, no work has investigated the effect of varying degrees of overlap between pretraining and CSR test instances for the state-of-the-art transformer-based models (BERT and RoBERTa). Any such investigation must include formulating a more precise definition of contamination.

Methods for purging easy instances from CSR benchmarks have been developed recently: for example, the algorithmic bias reduction of test sets proposed by \cite{sakaguchi2019winogrande} removes instances from the test set with exploitable annotation artifacts. These techniques depend on pre-computed neural network embeddings of a particular model, and so may be difficult for that model alone but not for previous or up-and-coming models. As the work of \newcite{zellers2018swag} and the follow-up by \newcite{zellers2019hellaswag} have shown, adversarial filtering must be iteratively re-adapted to newer models that may be immune to previous filtering. This may be costly. Adversarial filtering and related debiasing techniques also do not provide much insight on why certain test instances are filtered out. Our proposed method for data purging is interpretable and model-independent, and can be further supplemented with existing debiasing algorithms like AFLite \cite{sakaguchi2019winogrande} to ensure that benchmarks remain challenging.

\section{Hunting for Overlaps}

Our procedure for identifying train-test overlaps consists of three main steps: (1) parsing a test instance into its core components, (2) formulating a query using a schema derived from the parse, and (3) quantifying the degree of overlap between a train-test pair using an overlap scoring mechanism.

\vspace{-0.1cm}
\subsection{Skeletal Representation}

We first perform a partial parse of each test instance into a  general  skeleton  of  each  of  the  important  semantic  components,  in  the  order  that  they appear. We use rules related to the syntactic parse of the sentence implemented by Stanford CoreNLP \cite{manning2014corenlp}.

We use the notation in \newcite{emami-etal-2018-knowledge} to separate the components of WSC-like instances; that is, instances can be divided into a \textit{context} clause, which introduces the two competing antecedents, and a \textit{query} clause, which contains the target pronoun to be resolved:

\begin{align*}
& E_1, E_2   &  \text{the candidate antecedents} \\
& Pred_C     &  \text{the context predicate} \\
& +          &  \text{discourse connective} \\
& P          &  \text{the target pronoun} \\
& Pred_Q     &  \text{the query predicate} \\
\end{align*}

$E_1$ and $E_2$ are noun phrases in the context clause. In the WSC, these two are specified without ambiguity. $Pred_C$ is the context predicate  composed of the verb phrase that relates both antecedents to some event. The context contains $E_1$, $E_2$, and the context predicate $Pred_C$. The context and the query clauses are often connected by a discourse connective, $+$. The query contains the target pronoun, $P$, which is also specified unambiguously. Preceding or succeeding $P$ is the query predicate, $Pred_Q$, a verb phrase involving the target pronoun. In our case, we will treat  $Pred_C$ and $Pred_Q$ distinctly, and group all other components ($E_1$, $E_2$, $P$, $+$) together as \textit{content words} in the set $C$. Table~\ref{tab:examples} shows some examples of WSC instances and Table~\ref{tab:representation} shows sentence pairs in terms of each of these components.

\begin{table*}[ht]
\small 
\centering
\begin{tabular}{cL{13.5cm}} 
\hline

1 a) & The man couldn't lift his son because he was so \underline{weak}. (Answer: the man) \\
1 b) & The man couldn't lift his son because he was so \underline{heavy}. (Answer: son)\\

\hline

2 a) & The older students were bullying the younger ones, so we \underline{punished} them. (Answer: the older students)\\
2 b) & The older students were bullying the younger ones, so we \underline{rescued} them. (Answer: the younger ones)\\

\hline



3 a) & Sam tried to paint a picture of shepherds with sheep, but they ended up looking more like \underline{golfers}. (Answer: shepherds)\\
3 b) & Sam tried to paint a picture of shepherds with sheep, but they ended up looking more like \underline{dogs}. \newline (Answer: sheep)\\

\hline
\end{tabular}
\caption{Examples of Winograd instances. 
}
\label{tab:examples}

\vskip -.1in
\end{table*}

\begin{table*}[ht]
\small 
\begin{center}
\begin{tabular}{L{1.5cm}L{1.5cm}|L{2cm}L{2cm}L{0.8cm}L{1.3cm}} 
\toprule
 \centering Predicates & &  & \centering Content Words &  &  \\
\hline

 $Pred_C$ & $Pred_Q$ & $E_1$ & $E_2$ & $P$ & $+$ \\

\midrule

couldn’t lift & was so heavy & the man & his son & he & because \\

\hline

 were bullying & punished & the older students &the younger ones  & them & so \\

\hline

tried to paint & ended up looking more like & shepherds & sheep & they & but \\

\bottomrule
\end{tabular}
\caption{Winograd sentence pairs from Table~\ref{tab:examples}, parsed into the representation schema.}
\label{tab:representation}
\end{center}
\end{table*}
\vspace{-1cm}

\subsection{Query Schematization}
We use the above analysis of an instance to formulate a query used to retrieve similar instances in a text corpus. In particular, the per-instance query schema that we propose is:
\begin{center} $(\textit{Phrase}(\textit{Pred}_c, \textit{Pred}_Q,10)) \cap \bigcup\limits_{}^{C} c_{i}$,\\
\end{center}
where $\textit{Phrase}(\textit{Pred}_c,\textit{Pred}_Q,10)$ denotes that the two predicates must occur in the same order within a distance of 10 tokens to each other, and $c_i$ are content words in $C$ that may appear in any order in the sentence.

The choice of this schematization stems from the idea that the predicates are the most salient components of WSC-style problem instances. Instances of common-sense knowledge in corpora that support the resolution of a corresponding WSC instance often exhibit only these two components: for example, for 1 a) in Table~\ref{tab:examples}, a possible supporting instance is: \textit{John \underline{couldn't lift} Melissa and she \underline{was so heavy}}, although it only shares predicates (underlined) with 1 a). Nevertheless, content words may still contribute informatively and are included as optional components in the query. See Table~\ref{tab:query-examples} for the query extracted for the running example.

\begin{table*}[h]
\small 
\centering
\begin{tabular*}{\linewidth}{p{10cm}}
\toprule
\textbf{Sentence}: The man couldn't lift his son because he was so heavy. \\

\midrule

$Phrase$("couldn't lift", ``was so heavy", 10) $\cap$ (the man $\cup$ his son $\cup$ because)   \\ 

\bottomrule
\end{tabular*}
\caption{Query for an example Winograd sentence}
\label{tab:query-examples}
\vskip -.2in
\end{table*}

\subsection{Overlap scoring}
A retrieval function takes a query related to a given sentence, as formulated above, and estimates its relevance to a given document. In our case, ``documents'' correspond to individual sentences in the pretraining corpora. One popular retrieval function is BM25 \cite{Amati2009}, which is a bag-of-words-based function with various components and parameters.

Specifically, given a query $Q$ containing keywords $q_1,q_2…q_n$, the BM25 score of a document $D$ is:
\begin{equation*}
    \textit{score}(D,Q)=\sum_{i=1}^{n} \textit{IDF}(q_i)\cdot\frac{f(q_i,D)\cdot(k_1+1)}{f(q_i,D)+k_1 \cdot (1-b +b \cdot \frac{|D|}{\textit{avgdl}})},
\end{equation*}
where $\textit{IDF}(q_i)=log \frac{N-n(q_i)+0.5}{n(q_i)+0.5}$. Here, $f(q_i,D)$ is $q_i$’s term frequency in document $D$ in words, and $avgdl$ is the average document length in the text collection from which documents are drawn. Parameters $k_1$ and $b$ are free and usually chosen, in absence of a hyperoptimization, in the range [1.2, 2.0] and as 0.75, respectively.



We use the BM25 score in one of two ways:
\begin{enumerate}
    \item As a heuristic measure for the degree of overlap or relevance of a sentence in a pretraining corpus, with respect to a given test instance.
    \item As a cut-off criterion for sub-dividing a given CSR test set into overlapping and non-overlapping subsets.
\end{enumerate}

We use the Python package \textit{Whoosh} \cite{chaput2017whoosh}, which has methods to index pretraining corpora, to generate customized queries, and to score these based on the BM25 retrieval function. When the queries are customized in terms of logical operators as in our case, we employ a filter to remove sentences that do not meet the criteria. For example, a document that would otherwise have yielded a high relevancy score to a given query, but whose predicate words do not occur within the 10 token limit, would not be scored at all.

\begin{table*}[htp]
\small 
\begin{center}
\begin{tabular}{cL{10.5cm}} 
\hline

WSC Instance: & The man couldn't lift his son because he was so \underline{heavy}. Answer: son  \\
\hline 

Retrieved Sentences \& BM25 Scores: &``Nope , our driver had a steel plate in his back and couldn't lift anything ( although he was able to open the truck and put the rather heavy ramp in place , so I am not sure if he was unable or just lazy ) ."$~\rightarrow~$18.9\\
& ``Then I came across a box that weighed a ton - I couldn't even lift it it was so heavy$~\rightarrow~$26.5\\

 & ``1 man stopped to get it but he couldn't lift it because it was so heavy"$~\rightarrow~$36.1\\
& ``The man couldn't lift his son because he was so heavy"$~\rightarrow~$43.0 (exact copy)\\

\hline

WSC Instance: & Paul tried to call George on the phone, but he wasn't \underline{available}. Answer: George  \\
\hline 

Retrieved Sentences \& BM25 Scores: &``A couple of days later we tried to call him at home but his wife told us he wasn't available."$~\rightarrow~$25.3\\
& ``I also tried to call the district attorney, but, unsurprisingly, he wasn't available.$~\rightarrow~$32.9\\

 & ``Have a go and check if you're as intelligent as a human: Paul tried to call George on the phone, but he wasn't [successful/available]"$~\rightarrow~$33.4 (near exact copy)\\
& ``Paul tried to call George on the phone, but he wasn't available"$~\rightarrow~$40.3 (exact copy)\\

\hline

\end{tabular}
\caption{Example Sentences Retrieved and Corresponing BM25 scores for two WSC problems. 
}
\label{tab:ResolutionExamples}
\vspace{-0.5cm}
\end{center}
\end{table*}

In Table~\ref{tab:ResolutionExamples}, we provide examples of sentences retrieved from the pretraining corpora for a given test instance for various BM25 scores. Qualitatively, a trend appears towards an increased BM25 score with increasing relevance/degree of overlap between a test and pretraining instance. In the case where there was an exact copy found, the score was always significantly higher than when there was not. This suggests that these two steps of query schematization and BM25-based retrieval form an adequate (although by no means perfect) automatic heuristic for ranking the relevance and potential usefulness of pretraining instances.

\section{Experiments}

\subsection{Existing Benchmarks}

\paragraph{WSC \cite{levesque2011winograd}} The original pronoun disambiguation challenge, consisting of 273 problems. Each problem instance is manually crafted by an expert to avoid word association bias, although \newcite{trichelair2018evaluation} later report that 13.5\% of the questions may still exhibit such bias.

\paragraph{DPR \cite{rahman2012resolving}} DPR (Definite Pronoun
Resolution Dataset) provides 1,886 additional WSC-style problems authored by undergraduate students. \newcite{trichelair2018evaluation} observe that the dataset is likely less challenging than the original WSC because of an increased level of linguistic and dataset-specific biases. 
\paragraph{KnowRef \cite{emami-etal-2019-KnowRef}} KnowRef introduces over
8k WSC-style coreference resolution problems extracted and filtered using heuristic rules from 100 million web
sentences (from Reddit, Wikipedia, and OpenSubtitles).
\paragraph{Winogrande \cite{sakaguchi2019winogrande}} Winogrande is a large-scale dataset of 44k WSC-like problems, inspired by the original WSC but adjusted to improve both the scale and the difficulty of the dataset. The key steps of dataset construction are (1) a carefully designed crowdsourcing
procedure, followed by (2) systematic bias reduction against a finetuned RoBERTa model by adversarial filtering.

\subsection{Models}

\paragraph{BERT} BERT \cite{devlin2018bert} is a pretrained
neural language model with bidirectional paths and sentence representations in consecutive hidden layers. We
finetune BERT by splitting the input sentence into a context
and an option component using the candidate answer as delimiter as prescribed in \cite{devlin2018bert}.
We used grid-search for hyper-parameter tuning: learning
rate $\{1e - 5, 3e - 5, 5e -5\}$, number of epochs $\{3, 4, 5, 8\}$,
batch-size $\{8, 16\}$ with three different random seeds as in \cite{sakaguchi2019winogrande}. The pretraining corpora are BooksCorpus (800M words) \cite{zhu2015aligning} and English Wikipedia (2,500M words). We fine-tune BERT models on DPR-Train for the purpose of comparability with the state-of-the-art in \newcite{kocijan-etal-2019-surprisingly}, and we include that corpus as an additional source for querying overlaps.

\paragraph{RoBERTa} RoBERTa \cite{liu2019roberta} is an improved variant of BERT that adds more training data with larger batch
sizes and longer training, as well as other refinements like dynamic masking. RoBERTa performs consistently better
than BERT across many benchmarks. The pretraining corpora include those used for BERT and three more: CC-News \cite{nagel2016cc}, Openwebtext \cite{gokaslan2019openwebtext}, and the Stories Corpus \cite{trinh2018simple}.  We fine-tune RoBERTa models on the WNLI-train dataset \cite{wang2018glue} for comparability with the state-of-the-art model in \newcite{liu2019roberta}, including the corpus as an additional source for querying potential overlaps.

\subsection{Results}

In the following section, we report the performance of state-of-the-art models on subsets of the CSR test sets for which at least one overlapping instance was retrieved from the pretraining corpora -- that is, where the BM25 score between a train-test sentence pair is $>0$. In addition, we investigate how performance changes as we increase the BM25 score cut-off, and use this to assess the relationship between each test set as a whole and the pretraining corpora. Our motivation for these experiments is to gain insight as to the roles that increasingly relevant (and potentially duplicate) pretraining instances play in model performances on CSR benchmarks.

\begin{table*}[h]
\small
\centering
\begin{tabu}to \linewidth{@{}X[l,4.2]*3{X[1.5,c]}@{}}
\toprule
Test Set (size)                                &  WSC Test (273)  & DPR-Test (564)    & Winogrande-Dev (1,267)\\

 \midrule
\textbf{BERT, fine-tuned on DPR-Train} \cite{kocijan-etal-2019-surprisingly}
\\ \midrule
Overall Accuracy          & 71.4\%    & 84.9\% & 54.6 \% \\
\cmidrule(r){1-1}
Overlapping Subset Acc., BM25 $>$ 0 \\(subset size)                    & \textbf{79.2\%} (53) & \textbf{86.9\%} (246) & \textbf{59.1\%} (220)   \\ 
Non Overlapping Subset Acc., BM25 $=$ 0 (subset size)                   & 69.5\% (220) & 83.9\% (318) & 53.7\% (1,047)   \\ 
\midrule
\textbf{RoBERTa, fine-tuned on WNLI} \cite{liu2019roberta}
\\ \midrule
Overall Accuracy          & 91.2\%    & 93.1\% & 51.9 \% \\
\cmidrule(r){1-1}
Overlapping Subset Acc., BM25 $>$ 0 \\(subset size)                    & \textbf{91.7\%} (252) & \textbf{94.1\% }(306) & \textbf{55.8\%} (317)   \\ 
Non Overlapping Subset Acc., BM25 $=$ 0   \\(subset size)                 & 85.7\% (21) & 91.8\% (258) & 50.6\% (950)   \\

\cmidrule(r){1-1}

\end{tabu}

\caption{Performances of models  on test set subsets with and without potential overlaps\footnotemark[2] (BM25 score $>$0). Performances on overlapping subsets that corresponded to a significant difference (p$<$0.05) versus that on the non-overlapping subsets according to a chi-squared test are in bold.\\ }
\label{tab:performance}

\vspace{-0.8cm}
\end{table*}

\footnotetext[2]{Choice of fine-tuned models and training set come from the state-of-the-art in \newcite{kocijan-etal-2019-surprisingly} and \newcite{liu2019roberta}.}

Table~\ref{tab:performance} shows an increase in accuracy between 3\% and 10\% for both models on subsets of the test data for which overlap scores are greater than 0, suggesting that models tend towards better performance on instances for which there exist similar/overlapping pretraining instances.

\begin{figure}[h]
\centering

\includegraphics[width=0.7\linewidth,trim=4 4 4 4,clip]{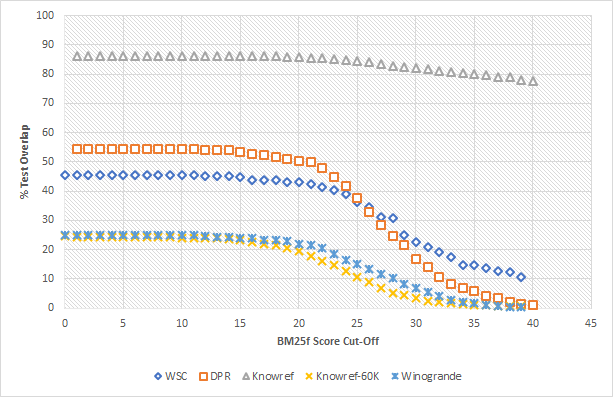}
\caption{\% Test set overlap as a function of BM25 score cut-off}
\label{fig:threshhold}

\vspace{-0.4cm}
\end{figure}

Next, we consider subsets of the test set corresponding to more stringent overlapping criteria, that is, for cut-offs of BM25 score $>25$ and $>35$ (Table~\ref{tab:cutoffs}). We chose these two cut-offs because they corresponded to sharp decreases in numbers of overlapping sentences retrieved\footnote[3]{While other choices are possible, the results do not vary dramatically within that window.} (see Figure~\ref{fig:threshhold}). An increased cut-off score reduces the size of the resulting subset substantially, but in many cases, significantly increased the performance difference. For the original WSC dataset, however, this trend did not hold when the cut-off increased considerably to $>35$; in fact, the performance difference dropped to negative. This may be explained by the fact that an exact copy of a test instance appearing in the pretraining corpus does not confer knowledge useful at test time, since these copies are by definition ambiguous. Consider retrieving \textit{The man could not lift the boy because he is so heavy.} This does not help resolution in the way that the less similar instance \textit{Tom could not lift Melissa because she was so heavy} does. Indeed, upon further investigation, we found that the WSC contains many exact overlaps with the pretraining corpora (26/29 of its instances with BM25$>$35 corresponded to exact copies.) This suggests that for CSR-based pronoun disambiguation tasks, current state-of-the-art models tend more towards the retrieval of highly relevant/similar sentences than they do towards memorizing exact (and otherwise unhelpful) duplicates of the test instances.

\begin{table*}[h]
\small
\centering
\begin{tabu}to \linewidth{@{}X[l,4.2]*3{X[1.5,c]}@{}}
\toprule
Test Set (size)                                &  WSC Test (273)  & DPR-Test (564)    & Winogrande-Dev (1,267)\\

 \midrule
\textbf{BERT, fine-tuned on DPR-Train} \cite{kocijan-etal-2019-surprisingly} & 71.4\%    & 84.9\% & 54.6 \% \\ 
\\ \midrule
Performance Difference (+)           \\
\cmidrule(r){1-1}
BM25 $>$ 0 (subset size)                   & \textbf{9.7\%} (53) & 2.2\% (246) & \textbf{5.4\%} (220)   \\ 
BM25 $>$ 25  (subset size)                   & \textbf{8.81\%}\% (29) & \textbf{4.46\%} (121) & \textbf{12.6\%} (92)   \\ 
BM25 $>$ 35  (subset size)                   & 28.9\% (6) & 8.6\% (15) & 5.4\% (9)   \\ 
\midrule
\textbf{RoBERTa, fine-tuned on WNLI}  \cite{liu2019roberta} & 91.2\%    & 93.1\% & 51.9 \%
\\ \midrule
Performance Difference (+)    \\      
\cmidrule(r){1-1}
BM25 $>$ 0 (subset size)                   & \textbf{6.0\%} (252) & 2.5\% (306) & \textbf{5.2\%} (317)   \\ 
BM25 $>$ 25  (subset size)                   & \textbf{5.3\%} (249) & \textbf{4.1\%} (186) & 0.82\% (190)   \\ 
BM25 $>$ 35  (subset size)                   & -2.6\% (236) & 7.3\% (32) & 4.1\% (25)   \\

\cmidrule(r){1-1}

\end{tabu}
\caption{Performance difference of models between overlapping and non-overlapping test sets for various BM25 cut-offs.\\ }
\label{tab:cutoffs}
\vspace{-0.9cm}
\end{table*}

Finally, we graphed the proportion of the original dataset with detected overlaps as a function of the BM25 score cut-off and use this to analyze the overlapping tendencies of each benchmark (Figure \ref{fig:threshhold}). Our findings demonstrate that a significant proportion of all datasets have overlaps receiving a BM25 score of at least 20 (ranging between 25\% to 68\%). We observe a decline as the threshold increases gradually to 40. In the case of most of these test sets, the decline asymptotically approaches 0; notably, however, WSC and KnowRef show no such trend. In the case of the former, the examination mentioned earlier yielded many examples where the test instance was referred to, \textit{verbatim}, in another context, often as a reference to the WSC itself. In the case of KnowRef, the pretraining corpus of Wikipedia was precisely the same corpus used to collect the KnowRef instances, yielding a significant number of  highly relevant overlaps. This suggests that various test sets may be subject to leakage/community overfitting (not surprising due to the notoriety and coverage the WSC has received) or contain test instances for which supporting knowledge may not be as long-tailed as anticipated.

\section{\corpusname{}}

\subsection{Limitations of previous test sets}

The results of the previous section suggest that overlapping test instances are, in general, less difficult for models to resolve, and this effect strengthens as the degree of overlap increases. In addition, current CSR test sets have a large proportion of overlapping instances with the pre-training corpora (BM25 Score $>$ 0), and while this proportion decreases as the degree of overlap increases, certain datasets have a considerable number of highly overlapping instances (BM25 Score $>$ 40). These results suggest two significant limitations of CSR benchmarks.

\paragraph{Highly Overlapping Instances:}

As shown in Figure~\ref{fig:threshhold}, WSC and KnowRef contain a considerable proportion of instances that overlap significantly with those in the pretraining corpora (either as exact copies or very similar sentences). In the case of KnowRef, almost all instances correspond closely to a sentence in English Wikipedia since this corpus was used specifically for data collection.



\paragraph{Predictable Structure:}
\vspace{-0.1cm}
A limitation common to WSC, DPR, and Winogrande is that they are based largely on the structural specifications of the WSC and some of its most famously cited instances (e.g. \textit{The trophy does not fit in the suitcase because it is too large}); that is, instances are often composed only of two clauses connected by a single causal discourse connective, like \textit{because}. For example, during the crowdsourcing protocol for Winogrande, annotators are first primed by classical examples of WSC sentences that may influence their creative process.
Test instances that are structurally similar to the original WSC instances seem more likely to overlap with training instances, and it is known that this kind of crowd-sourcing protocol engenders annotation artifacts \cite{gururangan2018annotation} that are particularly problematic when they do not corresponding to real-word data \cite{he2019unlearn}.

One can find more elaborate, real-word coreference examples that circumvent the above issues. For example, consider this sentence, taken \textit{as is} from Reddit: \\

\begin{exe}
\vspace{-0.4cm}
\item ``Forbes wrote that Edison can't be held accountable because his assistant willingly submitted to the trials and that the dangers of radiation poisoning were not well known."

\end{exe}

Here, despite the instance being a valid pronoun disambiguation problem akin to the WSC instances, there are multiple discourse connectives and more than two clauses, plus distractor content words that contribute variably to the correct resolution. All this renders it much more complex than most, if not all, of the WSC, DPR, and Winogrande instances. The idea that drives our corpus construction process is to identify and collect binary pronoun disambiguation problems as they occur naturally (and potentially, more intricately) in written text, while ensuring that the source of text is not contained in popular pretraining corpora.

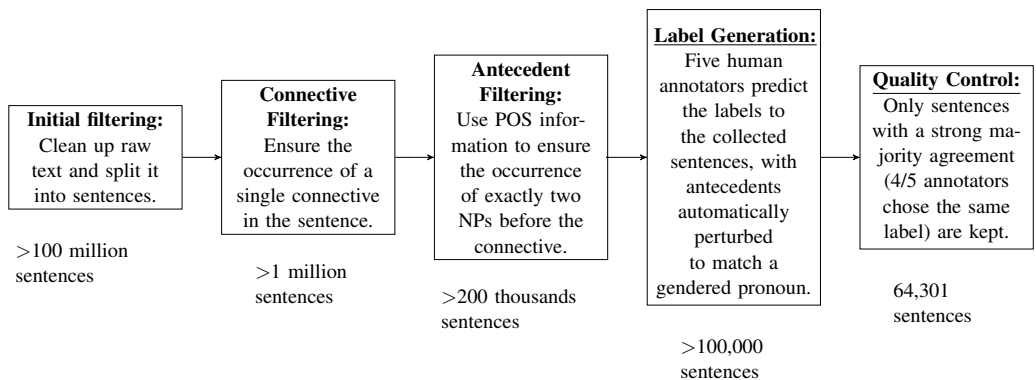
\begin {figure*}[htp]
\centering
\scalebox{0.70}{
 \begin{tikzpicture}
    [
    mynode/.style={rectangle, draw, align=center, text width =3cm ,minimum width=3cm, minimum height=1cm},
    widenode/.style={rectangle, draw, align=center, text width =4.5cm ,minimum width=4.5cm, minimum height=1cm}
]
    \node[mynode] (a) at ($(0, 0)$) {\textbf{Initial filtering:}\\ Clean up raw text and split it into sentences.};
    \node[mynode] (b) at ($(a) + (4, 0)$) {\textbf{Connective Filtering:}\\ Ensure the occurrence of a single connective in the sentence.};
    \node[mynode] (c) at ($(b) + (4, 0)$) {\textbf{Antecedent Filtering:}\\ Use POS information to ensure the occurrence of exactly two NPs before the connective.};
    \node[mynode] (d1) at ($(c) + (4, 0)$) {\textbf{\\\underline{Label Generation:}}\\ Five human annotators predict the labels to the collected \\sentences, with antecedents automatically perturbed to match a gendered pronoun.};
    \node[mynode] (e) at ($(c) + (8, 0)$) {\textbf{\underline{Quality Control:}}\\ Only sentences with a strong majority agreement (4/5 annotators chose the same label) are kept.};

    \draw (a) -> ($(a)!0.5!(b)$) edge[->] (b);
    \draw (b) -> ($(b)!0.5!(c)$) edge[->] (c);
    \draw (c) -| ($(c)!0.4!(d1)$) edge[->] (d1);
    \draw (d1) -| ($(d1)!0.55!(e)$) edge[->] (e);

    \node[below =5mm of a,text width=3cm,fill=white]
  (why1) {$>$100 million \\sentences};
  \node[below=5mm of b,text width=2cm,inner sep=.05cm,fill=white]
  (why2) {$>$1 million sentences};
  \node[below=5mm of c,text width=3cm,inner sep=.05cm,fill=white]
  (why3) {$>$200 thousands sentences};
  \node[below=5mm of e,text width=2cm,fill=white]
  (why4) {64,301 \\sentences};
  \node[below=0.5cm of d1,text width=2cm]
  (why4) {$>$100,000 sentences};

\end{tikzpicture}
}
\caption{\corpusname{}'s construction process. Differences from \newcite{emami-etal-2019-KnowRef} are underlined. }
\label{fig:corpcons}
\vspace{-0.5cm}
\end{figure*}

\subsection{Corpus Construction}

Motivated by the above limitations, we modify the corpus construction process of \newcite{emami-etal-2019-KnowRef} by scraping candidate sentences only from text documents not contained in the pretraining corpora of current models and using human annotators to resolve and label extracted sentences. Specifically, we scrape text samples from Reddit comments dating from 2006--2019. We filter this text through a multi-stage process to ensure quality and diversity as depicted in Figure \ref{fig:corpcons}, ultimately yielding 64,301 complex pronoun disambiguation problems scraped from written text. We compile and release these in a coreference task we call \corpusname{}. Examples of its instances are shown in Table~\ref{tab:examples2}.

\begin{table*}[htp]

\small 
\begin{center}
\begin{tabu}to\linewidth{@{}X[l]X[l,8]@{}} 
\toprule

Ex. 1: &\{Steven\} certainly manipulates \{Gregory\}, but [he] also has the best interest of the world at heart. (Steven) \\

\midrule

Ex. 2: & \{Gordon\} is a better baseball player than \{Joseph\}, but no one gives [him] any credit. (Gordon)\\

\midrule

Ex. 3: & \{Vernon\} was somewhat insulted and called \{Gary\} to see if [he] was goofing on him. (Gary) \\

\bottomrule
\end{tabu}
\caption{Examples of \corpusname{} instances. 
}
\label{tab:examples2}
\end{center}
\vspace{-0.7cm}
\end{table*}

\subsection{Overlap Statistics}

We perform experiments to tabulate overlap statistics for \corpusname{} and include the results in Figure~\ref{fig:threshhold} and Table~\ref{tab:performance2}. As seen in Figure~\ref{fig:threshhold}, \corpusname{}'s test set, without applying any explicit debiasing algorithm, yields the smallest proportion of test set overlaps. At the same time, it is nearly three times as large as the largest test set, Winogrande-dev. Table~\ref{tab:performance2} demonstrates also that, according to our retrieval technique, the subset of the test set with overlapping instances yields the lowest increase in performance (just above 2\%) for BERT and no change in performance for RoBERTa. Both BERT and RoBERTa achieve their lowest accuracy on the \corpusname{} test set (excepting Winogrande-dev, which was specifically debiased against these models to push their performance to chance).

\begin{table}[ht]
\small
\centering

\begin{tabu}to \linewidth{@{}X[l,0.2]*1{X[0.2,c]}@{}}
\toprule
                                &  \corpusname{}\\

 \midrule
\textbf{BERT, fine-tuned on DPR-Train} \cite{kocijan-etal-2019-surprisingly}
\\ \midrule
Overall Accuracy          & 67.9\%  \\
\cmidrule(r){1-1}
Overlapping Subset Acc., BM25 $>$ 0 (subset size)                    & 69.6\% (745)  \\ 
Non Overlapping Subset Acc., BM25 $=$ 0  (subset size)                   & 67.4\% (2,316) \\ 
\midrule
\textbf{RoBERTa, finetuned on WNLI} \cite{liu2019roberta}
\\ \midrule
Overall Accuracy          & 76.6\%    \\
\cmidrule(r){1-1}
Overlapping Subset Acc., BM25 $>$ 0 (subset size)                  & 76.1\% (770) \\ 
Non Overlapping Subset Acc., BM25 $=$ 0  (subset size)                     & 76.8\% (2,291)  \\

\cmidrule(r){1-1}

\end{tabu}
\captionsetup{justification=centering}
\caption{Performances of models on \corpusname{} with and without potential overlaps.\\ }
\label{tab:performance2}

\end{table}

\vspace{-0.5cm}

\section{Conclusion}

We proposed an automatic method of scoring the degree of overlap between test-train instances in CSR benchmarks and demonstrated that models generally perform better on test instances with high degrees of overlap in pretraining corpora. In response to our findings, we released a more difficult and largest-to-date WSC-style test set called \corpusname{}. We ensured that its overlaps with pretraining data are minimal by using a text source not contained in the suite of common pretraining corpora and by basing its construction on naturally-occurring sentences with no direct influence from WSC-like patterns. Our findings suggest that, for better or worse, highly similar pretraining instances have a significant influence on the performance of state-of-the-art transformer-based architectures. Coupled with the large fraction of exact copies or highly-overlapping instances that currently exist in CSR test sets, this effect may bias the evaluation and development of deep learning approaches for common-sense reasoning.


On a positive note, models still performed significantly better than random on non-overlapping test instances, and their relative rankings did not change. This suggests that the community's efforts do not seem to have overfit to the presence of overlaps yet: transformer-based language models still provide higher accuracy on the subset of non-overlapping instances as well as on our \corpusname{} dataset.

We therefore encourage researchers to be cognizant of such overlaps as important factors affecting the performance of CSR-based models, and to use this knowledge to have a clearer picture regarding the true capabilities of machine commonsense across these benchmarks. An important limitation of our analysis is that there is no guarantee that the overlapping subset is drawn from the same distribution as the original dataset, meaning that it is entirely possible that an emergent statistical bias (and neither retrieval nor memorization) caused the overlapping subset to be easier. Accordingly, our work also raises some open questions worthy of pursuit; firstly, how do we more precisely identify the cases of memorization/retrieval? Secondly, what constraints does the retrieval and/or memorization of models as a means of acquiring common sense present in terms of their capabilities/robustness?


\section*{Acknowledgements}
This work was supported by the Natural Sciences and Engineering Research Council of Canada and by Microsoft Research. Jackie Chi Kit Cheung is supported by the Canada CIFAR AI Chair program.

\bibliographystyle{acl}
\bibliography{coling2020}

\end{document}